\def\year{2020}\relax
\documentclass[letterpaper]{article} 
\usepackage{aaai20}  
\usepackage{times}  
\usepackage{helvet} 
\usepackage{courier}  
\usepackage[hyphens]{url}  
\usepackage{graphicx} 
\usepackage{graphics, amsmath}
\usepackage{color}
\usepackage{amsfonts}
\usepackage{amsthm}
\usepackage{amssymb}
\usepackage{multirow}
\usepackage{enumitem}
\usepackage{makecell}
\usepackage{fixltx2e}
\usepackage{bbm}
\usepackage{graphicx}  
\title{Select, Answer and Explain: Interpretable Multi-hop Reading Comprehension over Multiple Documents}
\author{Ming Tu, Kevin Huang, Guangtao Wang, Jing Huang, Xiaodong He, Bowen Zhou\\ 
JD AI Research\\ 
\{ming.tu, kevin.huang3, guangtao.wang, jing.huang, xiaodong.he, bowen.zhou\}@jd.com 
}
 
\begin{document}

\maketitle

\begin{abstract}
Interpretable multi-hop reading comprehension (RC) over multiple documents is a challenging problem because it demands 
reasoning over multiple information sources and explaining the answer prediction by providing 
supporting evidences. 
In this paper, we propose an effective and interpretable Select, Answer and Explain (SAE) system to solve the multi-document RC problem. 
Our system first filters out answer-unrelated documents and thus reduce the amount of distraction information. This is achieved by a document classifier trained with a novel pairwise learning-to-rank loss. 
The selected answer-related documents are then input to a model to jointly predict the answer and supporting sentences. 
The model is optimized with a multi-task learning objective on both token level for answer prediction and sentence level for supporting sentences prediction, together with an attention-based interaction between these two tasks. 
Evaluated on HotpotQA, a challenging multi-hop RC data set, the proposed SAE system achieves top competitive performance in distractor setting compared to other existing systems on the leaderboard.
\end{abstract}

\section{Introduction}
Machine Reading Comprehension (RC) or Question answering (QA) has seen great advancement in recent years. Numerous neural models have been proposed \cite{seo2016bidirectional,xiong2016dynamic,tay2018densely} and achieved promising performances on several different MRC data sets, such as SQuAD \cite{rajpurkar2016squad,rajpurkar2018know}, NarrativeQA \cite{kovcisky2018narrativeqa} and CoQA \cite{reddy2018coqa}. The performance was further boosted after the release of the Bidirectional Encoder Representations from Transformers (BERT) model \cite{devlin2018bert}, which has delivered state-of-the-art performance on several RC/QA data sets.

\begin{figure}[t]
\framebox{
\parbox{0.45\textwidth}{
\small
\textbf{Question:} What government position was held by the woman who portrayed Corliss Archer in the film Kiss and Tell? \smallskip\newline
\textbf{Document 1, Kiss and Tell (1945 film):} \textcolor{blue}{Kiss and Tell is a 1945 American comedy film starring then 17-year-old Shirley Temple as Corliss Archer.} In the film, two teenage girls cause their respective parents much concern when they start to become interested in boys. The parents' bickering about which girl is the worse influence causes more problems than it solves. \smallskip\newline
\textbf{Document 2, Shirley Temple:} \textcolor{blue}{Shirley Temple Black (April 23, 1928 - February 10, 2014) was an American actress, singer, dancer, businesswoman, and diplomat who was Hollywood's number one box-office draw as a child actress from 1935 to 1938. As an adult, she was named United States ambassador to Ghana and to Czechoslovakia and also served as Chief of Protocol of the United States.} \smallskip\newline
\textbf{Answer:} \textcolor{red}{Chief of Protocol} \smallskip\newline
\textbf{Supporting facts:} [``Kiss and Tell (1945 film)'',0], [``Shirley Temple'',0], [``Shirley Temple'',1]
}
}
\caption{An example from HotpotQA dev set (only 2 documents are shown). Supporting facts formatted as [document title, sentence id] are annotated in blue.} 
\label{fig:hotpotqa}
\end{figure}

Most existing research in machine RC/QA focuses on answering a question given a single document or paragraph. Although the performance on these types of tasks have been improved a lot over the last few years, the models used in these tasks still lack the ability to do reasoning across multiple documents when a single document is not enough to find the correct answer \cite{chen2019understanding}. In order to improve a machine's ability to do multi-hop reasoning over multiple documents, recently several multi-hop QA data sets have been proposed to promote the related research, such as \textsc{WikiHop} \cite{welbl2018constructing} and HotpotQA \cite{yang2018hotpotqa}. These data sets are challenging because they require models to be able to do multi-hop reasoning over multiple documents and under strong distraction. HotpotQA also encourages explainable QA models by providing supporting sentences for the answer, which usually come from several documents (a document is called "gold doc" if it contains the answer or it contains supporting sentences to the answer).
One example from HotpotQA is shown in Figure \ref{fig:hotpotqa}. Besides the answer ``Chief of Protocol'', HotpotQA also annotates supporting sentences (text in blue) in gold documents to explain the choice of that answer.

To solve the multi-hop multi-document QA task, two research directions have been explored. The first direction focuses on applying or adapting previous techniques that are successful in single-document QA tasks to multi-document QA tasks, for example the studies in \cite{dhingra2018neural,zhong2019coarse,yang2018hotpotqa,nishida2019answering}. The other direction resorts to Graph Neural Networks (GNN) to realize multi-hop reasoning across multiple documents, and promising performance has been achieved \cite{song2018exploring,de2018question,cao2019bag,tu2019hdegraph,xiao2019dfgn}.



Despite of the above achieved success, there are still several limitations of the current approaches on multi-hop multi-document QA. First, little attention has been paid to the explainability of the answer prediction, which is important and challenging because in multi-hop multi-documents QA tasks supporting evidences could spread out over very long context or multiple documents \cite{welbl2018constructing}. Being able to retrieve evidences from scattering information in different context results in more practical and intelligent QA systems. 

Second, almost all existing methods directly work on all documents either by concatenating them or processing them separately, regardless of the fact that most context is not related to the question or not helpful in finding the answer. Few attempts have been conducted to design a document filter in order to remove answer-unrelated documents and reduce the amount of information needs to be processed. An accurate document selection module can also improve the scalability of a QA model without degradation on performance \cite{min2018efficient}. 

Third, current applications of GNN for QA tasks usually take entities \cite{de2018question,tu2019hdegraph,xiao2019dfgn} as graph nodes and reasoning is achieved by conducting message passing over nodes with contextual information. This is only possible when a predefined set of target entities is available (e.g. \cite{welbl2018constructing}). Otherwise a Named Entity Recognition (NER) tool is used to extract entities, which could result in redundant and noisy entities for graph reasoning. If the answer is not a named entity, further processing is needed to locate the final answer \cite{xiao2019dfgn}.

To overcome these limitations, we propose an effective and interpretable system called Select, Answer and Explain (SAE) to solve the multi-hop RC over multiple documents. 
Our SAE system first filters out answer-unrelated documents and thus reduce the amount of distraction information. This is achieved by a document classifier trained with a novel pairwise learning-to-rank loss, which can achieve both high accuracy and recall on the gold documents.
The selected gold documents are then input to a model to jointly predict the answer to the question and supporting sentences.
The model is optimized in a multi-task learning way on both token level for answer span prediction and sentence level for supporting sentence prediction. While the answer prediction is accomplished by sequential labeling with start and end tokens as targets, we cast the support sentence prediction as a node classification task. We build a GNN model to do reasoning over contextual sentence embeddings, which is summarized over token representations based on a novel mixed attentive pooling mechanism. Multi-task learning together with the mixed attention-based interaction between these two tasks ensures that complementary information between the two tasks is exploited. 

Our proposed SAE system is evaluated on the distractor setting of the HotpotQA data set.
On the blind test set of HotpotQA, our SAE system attains top competitive results compared to other systems on the leaderboard \footnote{https://hotpotqa.github.io/} at the time of submission (Sep 5th).
To summarize, we make the following contributions:
\begin{enumerate}
    \item We design a document selection module to filter out answer-unrelated documents and remove distracting information. The model is based on multi-head self-attention over document embeddings to account for interaction among documents. We propose a novel pairwise learning-to-rank loss to further enhance the ability to locate accurate top ranked answer-related gold documents.
    \item Our ``answer and explain'' model is trained in a multi-task learning way to jointly predict answer and supporting sentences on gold documents. We build multi-hop reasoning graphs based on GNN with contextual sentence embeddings as nodes, instead of using entities as nodes as in previous work, to directly output supporting sentences with the answer prediction. 
    \item The contextual sentence embedding used in GNN is summarized over token representations based on a novel mixed attentive pooling mechanism. The attention weight 
    is calculated from both answer span logits and self-attention output on token representations. This attention based interaction enables exploitation of complementary information between ``answer'' and ``explain'' tasks.
\end{enumerate}

\section{Related work}

\textbf{Multi-hop multi-document QA:} The work in \cite{dhingra2018neural} designed a recurrent layer to explicitly exploit the skip connections between entities from different documents given coreference predictions. An attention based system was proposed in \cite{zhong2019coarse} and it shows that techniques like co-attention and self-attention widely employed in single-document RC tasks are also useful in multi-document RC tasks.
The study by \cite{song2018exploring} adopted two separate Named Entity Recognition (NER) and coreference resolution systems to locate entities in support documents, which are then used in GNN to enable multi-hop reasoning across documents. Work by \cite{de2018question,cao2019bag} directly used mentions of candidates as GNN nodes and calculated classification scores over mentions of candidates. The study in \cite{tu2019hdegraph} proposed a heterogeneous graph including document, candidate and entity nodes to enable rich information interaction at different granularity levels.

Our proposed system is different from the previous models in that 1) Our model is jointly trained and is capable of explaining the answer prediction by providing supporting sentences.  2) We propose to first filter out answer unrelated documents and then perform answer prediction.

\noindent \textbf{Explainable QA:} The study in \cite{zhou2018interpretable} proposed an Interpretable Reasoning Network for QA on knowledge base. The baseline model provided in the HotpotQA paper \cite{yang2018hotpotqa} and the QFE model proposed in \cite{nishida2019answering} are based on a single-document RC system proposed in \cite{clark2018simple}, with interpretable answer prediction. However multi-hop reasoning was not explicitly dealt with in this work. The DFGN model proposed in \cite{xiao2019dfgn} considered the model explainability by first locating supporting entities and then leading to support sentences, while our model directly finds the supporting sentences. The study in \cite{wang2019evidence} designed an evidence extractor for QA tasks based on Deep Probabilistic Logic, while our model is based on GNN by taking the interaction among sentences within or across documents into account.

\noindent \textbf{Sentence or document selector}: The study in \cite{min2018efficient} proposed an efficient sentence selector for single-document QA tasks, while our selector works on multiple documents. The DFGN model \cite{xiao2019dfgn} also has a document selector however they did not consider the relational information among documents and treated each document independently.

\section{Methodology}
\begin{figure}[t]
    \centering
    \includegraphics[width=0.6\linewidth]{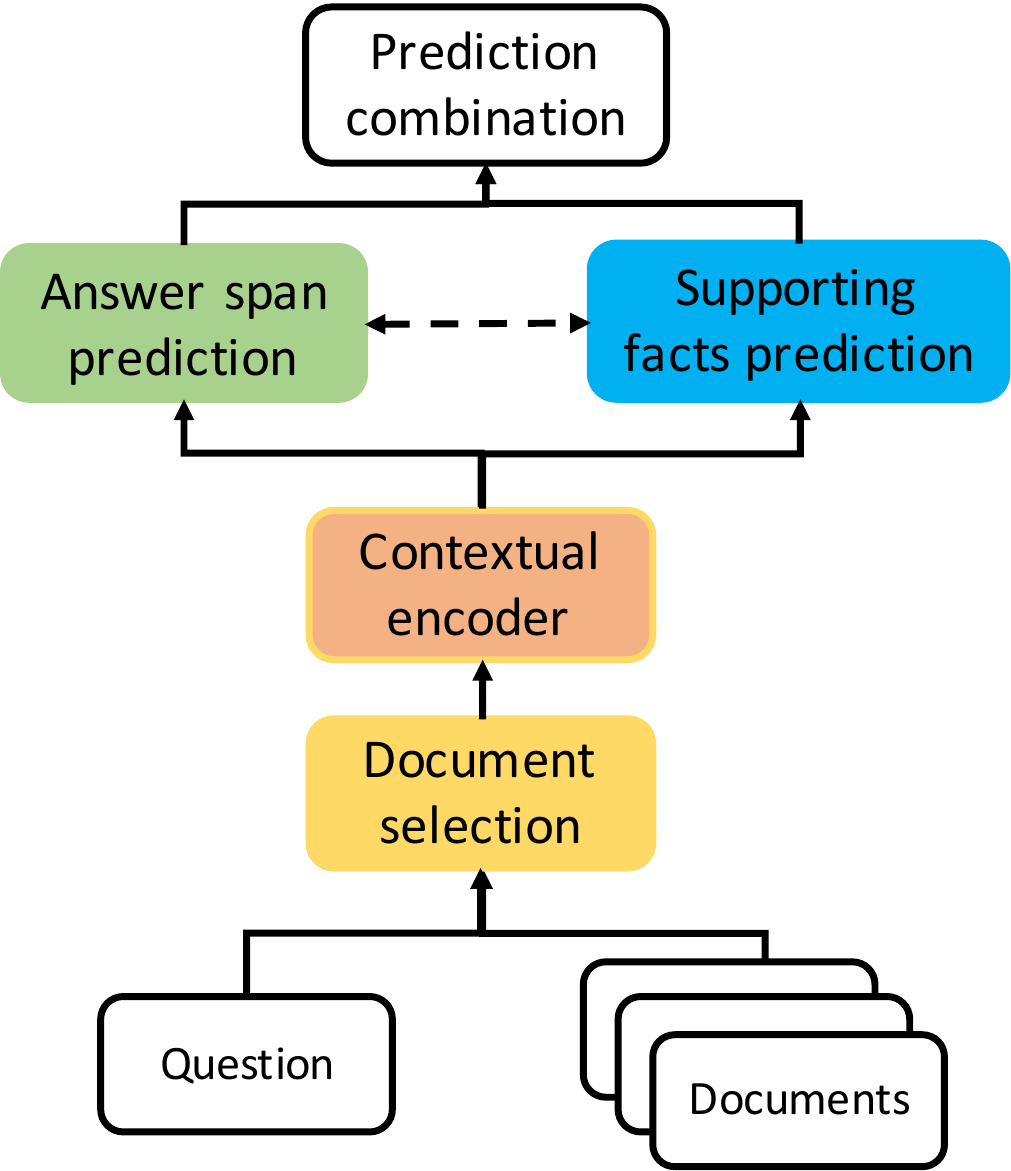}
    \caption{Diagram of the proposed SAE system. The dashed arrow line indicates the mixed attention based interaction between the two tasks}
    \label{fig:diagram}
\end{figure}

The diagram of the proposed SAE system is shown in Figure \ref{fig:diagram}. We assume a setting where each example in our data set contains a question and a set of $N$ documents; a set of labelled support sentences from different documents; the answer text, which could be a span of text or ``Yes/No''. 
We derive the gold document labels from the answer and support sentence labels. We use $D_{i}$ to note document $i$: it is labelled as $1$ if $D_i$ is a gold doc, otherwise $0$. We also label the answer type as one of the following annotations: ``Span'', ``Yes'' and ``No''. 

\subsection{\textbf{S}elect gold documents}
\label{select_gold}

\begin{figure}[t]
    \centering
    \includegraphics[width=0.9\linewidth]{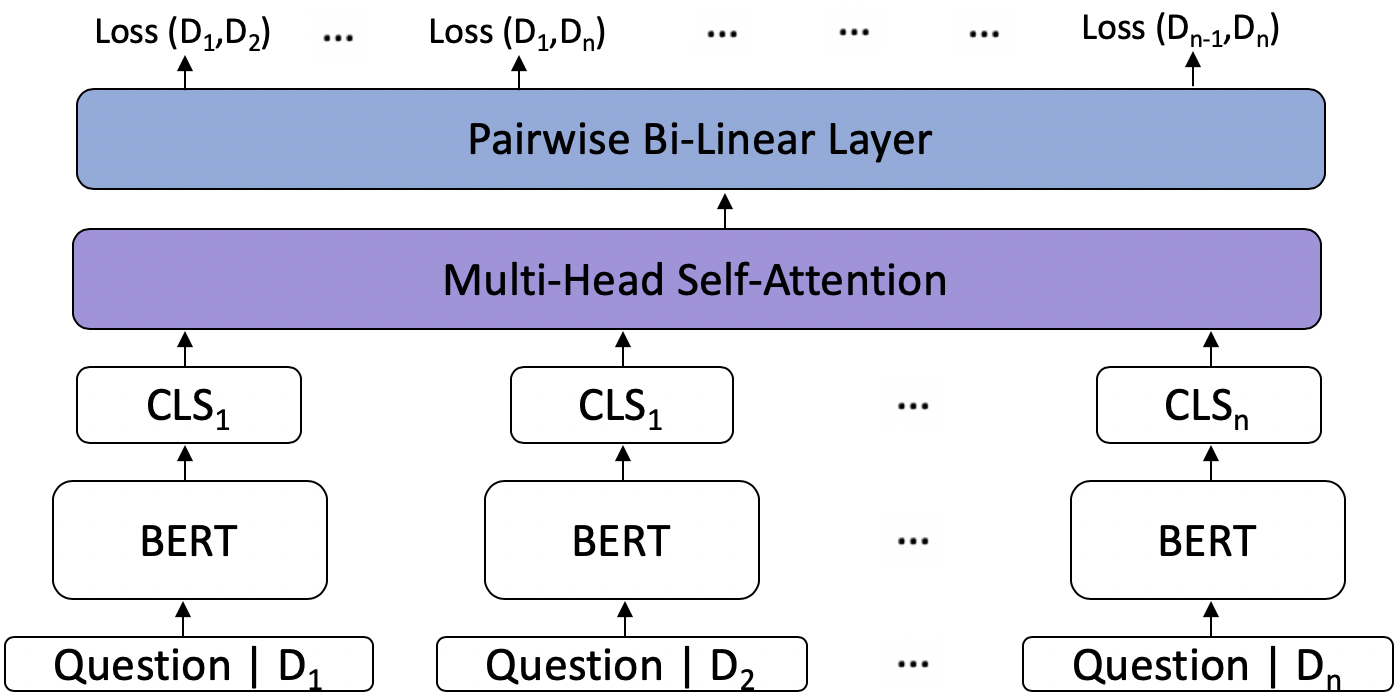}
    \caption{Diagram of document selection module. $N$ indicates the total number of documents.}
    \label{fig:extractor}
\end{figure}

Our SAE system begins by accurately extracting gold documents from the given input of $N$ documents. This phase of our system is crucial in minimizing distracting information being passed to the downstream answer prediction and explanation generation tasks. For every document, we generate an input to feed through BERT by concatenating ``[CLS]'' + question + ``[SEP]'' + document + ``[SEP]''. We use the ``[CLS]'' token output from BERT as a summary vector for each question/document pair. A straightforward way for gold document classification is to project this summary vector into a singular dimension and calculate the binary cross entropy loss as in \cite{xiao2019dfgn}:

\begingroup\makeatletter\def\f@size{8}\check@mathfonts
\def\maketag@@@#1{\hbox{\m@th\large\normalfont#1}}%
\begin{align}
    L = -\sum_{i=0}^{n} t_{i}logP(D_{i}) + (1-t_{i})log(1-P(D_{i}))
\end{align}\endgroup
where $t_{i}$ is the label of $D_i$, $n$ is the number of documents, and $P(D_{i})$ is the probability of document $i$ being in label $t_i$. 
This simple approach treats each document separately, without considering inter-document interactions and relationships that are essential for the downstream multi-hop reasoning task.

Therefore we propose a new model as shown in Figure \ref{fig:extractor}. First we add a multi-head self-attention (MHSA) layer on top of the ``CLS'' tokens. This MHSA layer is defined as:
\begin{align}
    & Attention = softmax(\frac{QK^{T}}{\sqrt{d_{k}}}) \\
    & Multihead = Concat(head_{i}...head_{n})W^{o}  \\
    & head_{i} = Attention(QW_{i}^{Q},KW_{i}^{k},VW_{i}^{v})
\end{align}
where, $Q$, $K$, and $V$ are linear projections from ``CLS'' embeddings of documents, representing attention queries, key and values \cite{vaswani2017attention}. Our motivation for adding attention across the ``CLS'' tokens generated from different documents is to encourage inter-document interactions. Inter-document interactions are crucial to 
the multi-hop reasoning across documents. 
By allowing document/question representations to interact with each other, the model is able to train on a better input signal for selecting a set of gold documents that are needed for the answer extraction and support sentence prediction. 


We also formulate our problem from a classification problem to a pairwise learning-to-rank problem, where we would take the top ranked documents as predicted gold documents \cite{liu2009learning}. 
By comparing a document to all other documents, the model is able to better distinguish a small set of gold documents from the rest distracting documents.
We first give a score $S(.)$ to each document $D_{i}$: $S(D_{i}) = 1$ if $D_{i}$ is a golden document, and $S(D_{i}) = 0$ otherwise. 
We then label each pair of input documents: given a pair of input documents $(D_{i},D_{j})$, our label $l$ will be set as:
\[ 
   l_{i,j}=
   \begin{cases} 
       1 & \text{if $S(D_{i}) > S(D_{j})$}\\
       0 & \text{if $S(D_{i}) <= S(D_{j})$}  \\
   \end{cases}
\]
This label indicates that gold document would be scored higher than non-gold docs. 

We also consider the document containing the answer span to be more important for downstream tasks. Thus we give $S(D_{i}) = 2$ if $D_{i}$ is a golden document containing the answer span. 

Our model outputs a probability for each pair of documents by passing the MHSA outputs of documents through a bi-linear layer, which is trained on binary cross entropy as the following:
\begingroup\makeatletter\def\f@size{7}\check@mathfonts
\def\maketag@@@#1{\hbox{\m@th\large\normalfont#1}}%
\begin{equation}
    L = -\sum_{i=0}^{n}\sum_{j=0,j \neq i}^{i} l_{i,j}logP(D_{i},D_{j}) + \\ (1-l_{i,j})log(1-P(D_{i},D_{j}))
\end{equation}
\endgroup
where $l_{i,j}$ is the label for the pair of documents $(D_{i},D_{j})$. $P(D_{i},D_{j})$ is our model's predicted probability that $D_{i}$ is more relevant than $D_{j}$. At inference time, we examine the comparisons that each of the documents makes with each other. We threshold each of the pair of predictions with a value of 0.5, and define relevance as $R_{i} = \sum_{j}^{n} \mathbbm{1}(P(D_{i},D_{j}) > 0.5)$. $\mathbbm{1}$ denotes the indicator function. We take the top $k$ ranked documents from this relevance ranking as our filtered documents.

\subsection{Answer and Explain}
The document selection module removes answer unrelated documents and distractions from the original input. 
Given the question and gold documents, 
we jointly train the answer prediction and supporting sentence classification in a multi-task learning way (note that at inference time we use the predicted gold documents.). In addition we explicitly model the interaction between the two tasks with attention-based summarized sentence embeddings.

First, all the gold documents are concatenated into one context input, and BERT is employed to encode the question and context pair in ``[CLS] + question + [SEP] + context + [SEP]'' format. Denote the token-level BERT output for $i$th input as $\mathbf{H}^i=\{\mathbf{h}^i_0, \mathbf{h}^i_2, \cdots, \mathbf{h}^i_{L-1}\}$, and we expect $\mathbf{H}^i \in \mathbb{R}^{L \times d}$ to have enough question-related contextual information, $d$ is the output dimension of BERT. Next, we introduce our new design of token-level and sentence-level multi-task learning.

\subsubsection{Answer prediction}
A 2-layer Multilayer Perceptron (MLP) with output size 2 is applied to the BERT output $\mathbf{H}^i$. One dimension of the output is used for start position prediction, and the other for end position prediction, as introduced in \cite{devlin2018bert}. We calculate the cross entropy between logits of all possible indices from $1$ to $L$, and the true start and end position of answer span. The final output is the predicted answer span derived from the start and end positions. The process could be described using the following formulas.

\begin{equation}
    \mathbf{\hat{Y}} = f_{span}(\mathbf{H}^i) \in \mathbb{R}^{L \times 2},
\end{equation}
\begin{equation}
    L^{span} = \frac{1}{2}(CE(\mathbf{\hat{Y}}[:,0], \mathbf{y}^{start}) + CE(\mathbf{\hat{Y}}[:,1], \mathbf{y}^{end})),
\end{equation}
where the first row of $\mathbf{\hat{Y}}$ is the logits of start position and second row is the logits of end position. $\mathbf{y}^{start}$ and $\mathbf{y}^{end}$ are the labels of start and end positions in the range [0, L-1]. $CE$ denotes cross entropy loss function.

\subsubsection{Supporting sentence prediction}
The other task is to predict whether a sentence in the input context is a supporting evidence to the answer prediction. To achieve sentence-level prediction, we first obtain sequential representation of each sentence from $\mathbf{H}^i$. 
\begin{equation}
     \mathbf{S}^j = \mathbf{H}[j^{s}:j^{e},:] \in \mathbb{R}^{L^j \times d},
\end{equation}
where $\mathbf{S}^j$ is the matrix representing the token embeddings within sentence $j$ (for clarity, we drop the sample index $i$ here); $j^s$ and $j^e$ defines the start and end positions, and $L^j$ is the length of sentence $j$.

Intuitively, the answer prediction task and support sentence prediction task could complement each other. Our observation is that the answer prediction task could always help support sentence prediction task because the sentence with answer is always a piece of evidence; however it is not the same case the other way around because there may be multiple support sentences and the sentence with the highest probability may not contain the answer. Therefore, to exploit the interaction between the two complementary tasks, we propose an attention-based summarized sentence representation to introduce complementary information from answer prediction. The attention weights  are calculated in the following way: one part of attention is computed with self-attention on $\mathbf{S}^j$; the other part is from the summation of the start and end position logits from the answer prediction task.
\begin{equation}
    \boldsymbol{\alpha^j} = \sigma(f_{att}(\mathbf{S}^j) + \mathbf{\hat{Y}}[j^{s}:j^{e},0] + \mathbf{\hat{Y}}[j^{s}:j^{e},1]),
    \label{equ:mixed_att}
\end{equation}
\begin{equation}
    \mathbf{s}^j = \sum_{k=0}^{L^j} \alpha^j_k \mathbf{S}^j[k,:] \in \mathbb{R}^{1 \times d},
    \label{equ:sent_att}
\end{equation}

where $f_{att}$ is a two-layer MLP with output size 1, and function $\sigma$ is the softmax function applied on the sequence length dimension. $\boldsymbol{\alpha}^j \in \mathbb{R}^{L^j \times 1}$ denotes the attention weight on each token of sentence $j$. Our mixed attention idea is different from the supervised attention proposed in \cite{rei2019jointly} because we calculate attention weights from two inputs while supervised attention only uses token-level logits for attention weight calculation.

\begin{figure}[t]
    \centering
    \includegraphics[width=0.5\linewidth]{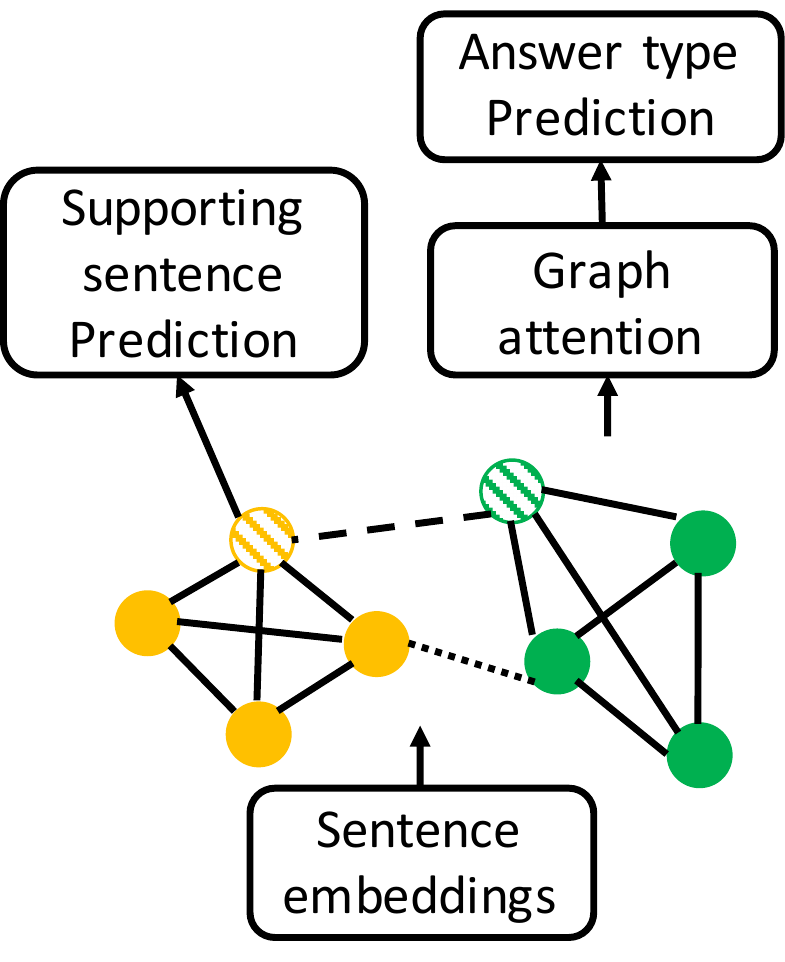}
    \caption{Diagram of the supporting sentence prediction module. Different node colors indicate that nodes are from different documents. Nodes with diagonal stripes indicate that the corresponding input sentences are supporting sentences. We use solid, dashed and dotted lines to differentiate different edge types.}
    \label{fig:e_module}
\end{figure}

Next we propose to build a GNN model over sentence embeddings $\mathbf{s}^j$ to explicitly facilitate multi-hop reasoning over all sentences from the predicted gold documents for better exploitation of the complex relational information. We illustrate our model for support sentence prediction in Figure \ref{fig:e_module}. We use sentence embedding $\mathbf{s}^j$ to initialize the graph node features. A multi-relational Graph Convolution Networks (GCN) based message passing strategy is employed to update the graph node features, and the final node features are input to a MLP to get the classification logit of each sentence.

To build the connections among graph nodes, we design three types of edges based on the named entities and noun phrases presented in the question and sentences:
\begin{enumerate}
\item Add an edge between two nodes if they are originally from the same document (solid line in Figure \ref{fig:e_module}).
\item Add an edge between two nodes from different documents if the sentences representing the two nodes both have named entities or noun phrases (can be different) in the question (dashed line in Figure \ref{fig:e_module}).
\item Add an edge between two nodes from different documents if the sentences representing the two nodes have the same named entities or noun phrases (dotted line in Figure \ref{fig:e_module}).
\end{enumerate}
The motivation for the first type of edge is that we want the GNN to grasp the global information presented within each document. Furthermore, cross-document reasoning is achieved by jumping from entities in the question to unknown bridging entities or comparing attributes of two entities in the question \cite{yang2018hotpotqa}. Thus, we design the second and third types of edge to better capture such cross-document reasoning path.

For message passing, we use multi-relational 
GCN with gating mechanism as in \cite{de2018question,tu2019hdegraph}. Assume $\mathbf{h}_j^0$ represents initial node embedding from sentence embedding $\mathbf{s}_j$, the calculation of node embedding after one hop (or layer) can be formulated as 
\begin{equation}
    \mathbf{h}_j^{k+1} = act(\mathbf{u}_j ^ k) \odot \mathbf{g}_j ^ k + \mathbf{h}_j ^ k \odot (1-\mathbf{g}_j ^ k)
\end{equation}
where
\begin{equation}
    \mathbf{u}_j ^ k = f_s(\mathbf{h}_j ^ k) + \sum_{r \in \mathcal{R}} \frac{1}{\left | \mathcal{N}_j ^ r \right |} \sum_{n \in \mathcal{N}_j ^ r} f_r(\mathbf{h}_n ^ k),
\end{equation}
\begin{equation}
    \mathbf{g}_j ^ k = sigmoid(f_g([\mathbf{u}_j ^ k; \mathbf{h}_j ^ k])).
\end{equation}
$\mathcal{R}$ is the set of all edge types, $\mathcal{N}_j ^ r$ is the neighbors of node $j$ with edge type $r$ and $\mathbf{h}_n ^ k$ is the node representation of node $n$ in layer $k$. $\left | \cdot \right|$ indicates the size of the neighboring set. 

Each of $f_r$, $f_s$, $f_g$ defines a transform on the input node representations, and can be implemented with a MLP. Gate control $\mathbf{g}_j ^ k$, which is a vector consisting of values between 0 and 1, is to control the amount information from computed update $\mathbf{u}_j ^ k$ or from the original node representation $\mathbf{h}_j ^ k$. Function $act$ denotes a non-linear activation function. After message passing on the graph with predefined number of hops, each node has its final representation $\mathbf{h}_j$. We use a two-layer MLP $f_{sp}$ with 1-dimensional output to get the logit of node $j$ for supporting sentence prediction: $\hat{y}_j^{sp} = sigmoid(f_{sp}(\mathbf{h}_j)$).

Besides the supporting sentence prediction task, we add another task on top of the GNN outputs to account for ``Yes/No'' type of questions. We formulate the answer type prediction task as a 3-class (``Yes'', ``No'' and ``Span'') classification. We design a simple graph attention module to calculate the weighted sum of all nodes representation over $j$ by $\mathbf{h} = \sum_j a_j \mathbf{h}_j$, where the attention weight $\mathbf{a}$ is derived by $\mathbf{a} = \sigma(\hat{\mathbf{y}}^{sp})$.
Then, another two-layer MLP with 3-dimensional output is applied to the weighted-summed graph representation to get the answer type prediction: $\hat{\mathbf{y}}^{ans} = f_{ans}(\mathbf{h})$.
  
The final training loss is the summation of span prediction, support sentence prediction and answer type prediction losses:
\begin{equation}
    L = \gamma L^{span} + BCE(\hat{\mathbf{y}}^{sp}, \mathbf{y}^{sp}) + CE(\hat{\mathbf{y}}^{ans}, \mathbf{y}^{ans})
\end{equation}
where $BCE()$ represents binary cross entropy loss function; $\mathbf{y}^{sp}$ and $\mathbf{y}^{ans}$ are the labels of support sentences and answer types respectively. We add a weight $\gamma$ to span loss to account for the scale difference of different losses.

\section{Experiments}

\begin{table*}[]
    \centering
    \caption{Results comparison between our proposed SAE system with other methods. $^*$ indicates unpublished models.}
    \resizebox{0.65\linewidth}{!}{
    \begin{tabular}{c|c|cc|cc|cc}
    \hline
        \multirow{2}*{}&\multirow{2}*{Model} & \multicolumn{2}{c|}{Ans} & \multicolumn{2}{c|}{Sup} &         \multicolumn{2}{c}{Joint} \\
        ~ & ~& EM & $F_1$  & EM & $F_1$  & EM & $F_1$ \\
        \hline
        \multirow{4}*{Dev}&~Baseline\cite{yang2018hotpotqa} & 44.44 & 58.28 & 21.95 & 66.66 & 11.56 & 40.86\\
        ~&QFE\cite{nishida2019answering}& 53.70 & 68.70 & 58.80 & 84.70 & 35.40 & 60.60 \\
        ~&DFGN\cite{xiao2019dfgn}& 55.66 & 69.34 & 53.10 & 82.24 & 33.68 & 59.86 \\
        ~&SAE(ours)&61.32 & 74.81 & 58.06 & 85.27 & 39.89 & 66.45\\
        ~&SAE-oracle(ours)&63.48 & 77.16 & 62.80 & 89.29 & 42.77 & 70.13 \\
        ~&SAE-large(ours)&67.70 & 80.75 & 63.30 & 87.38 & 46.81 & 72.75 \\
        \hline
        \hline
        \multirow{5}*{Test}&~Baseline\cite{yang2018hotpotqa} & 45.46 & 58.99 & 22.24 & 66.62 & 12.04 & 41.37 \\
        ~&QFE\cite{nishida2019answering}& 53.86 & 68.06 & 57.75 & 84.49 & 34.63 & 59.61 \\
        ~&DFGN\cite{xiao2019dfgn} & 56.31 & 69.69 & 51.50 & 81.62 & 33.62 & 59.82\\
        ~&SAE(ours) & 60.36 & 73.58 & 56.93 & 84.63 & 38.81 & 64.96 \\ 
        ~&SAE-large(ours) & 66.92 & 79.62 & \textbf{61.53} & 86.86 & \textbf{45.36} & 71.45 \\ \cline{2-8}
        ~&C2F Reader$^*$ & 67.98 & 81.24 & 60.81 & 87.63 & 44.67 & 72.73 \\
    \hline
    \end{tabular}}
    \addtolength{\belowcaptionskip}{-12pt}
    \label{tab:result}
\end{table*}

\subsection{Data set}
HotpotQA is the first multi-hop QA data set taking the explanation ability of models into account. HotpotQA is constructed in the way that crowd workers are presented with multiple documents and are asked to provide a question, corresponding answer and support sentences used to reach the answer as shown in Figure \ref{fig:hotpotqa}. The gold documents annotation can be derived from the support sentences annotation. There are about 90K training samples, 7.4K development samples, and 7.4K test samples. Please refer to the original paper \cite{yang2018hotpotqa} for more details.

HotpotQA presents two tasks: answer span prediction and supporting facts prediction. Models are evaluated based on Exact Match (EM) and $F_1$ score of the two tasks. Joint EM and $F_1$ scores are used as the overall performance measurements, which encourage the model to be accurate on both tasks for each example. In the experiment, we train our SAE system on the training set, and tune hyperparameters on the development set. Our best model on development set is submitted to the leaderboard organizer to obtain the performance measurements on the blind test set.

\subsection{Implementation details}

Our implementation is based on the PyTorch \cite{paszke2017automatic} implementation of BERT\footnote{https://github.com/huggingface/pytorch-pretrained-BERT}.  We use both \textit{BERT base uncased} model (``SAE'') and Roberta \cite{liu2019roberta} large model (``SAE-large''). All texts are tokenized using the wordpiece tokenizer adopted by BERT and Roberta. Since only answer text is provided in HotpotQA, we employ a simple exact match strategy to find the start and end tokens of the answer in documents, and use it as the answer prediction labels. We also use spaCy\footnote{https://spacy.io} to recognize named entities and noun phrases in context and question. 

To train the document selection module, we use all samples in the training set. 
For the multi-task learning module, we only use the annotated gold documents for training and evaluate the model on predicted gold documents. 
During evaluation, we only take the top-2 documents returned by the document selection module as input to the answer and support sentence prediction module. if the answer type prediction is ``Span'', the answer span output is used as predicted answer; otherwise, the predicted answer is ``Yes'' or ``No'' based on the answer type prediction.

\subsection{Results}

In Table \ref{tab:result}, we show the performance comparison among different models on both development and blind test set of HotpotQA. On development set, our method improves more than 28\% and 25\% absolutely in terms of joint EM and $F_1$ scores over the baseline model. Compared to the DFGN model \cite{xiao2019dfgn} and QFE model \cite{nishida2019answering}, our SAE system is over 5\% absolutely better in terms of both joint EM and $F_1$ scores. 

We also show the results of SAE-oracle and SAE-large here. For SAE-oracle, we directly input the annotated gold documents of dev set to get answer and support sentence prediction. We find that for almost all measurements, the gap between our model using predicted gold documents and oracle gold documents is around 3-4\%, which implies the effectiveness of our document selection module. By using large pre-trained language models as encoders, the performance is improved tremendously. 

On the blind test set, since most submitted systems on the HotpotQA leaderboard are unpublished, we only show the comparison with two published models, and the best single model on the leaderboard. Our SAE-large model ranks No. 2 at the time of submission.
Compared to the best single model, our SAE system performs better in terms of support sentence prediction EM and joint EM.

\subsection{Ablation studies}

\begin{table}[]
\centering
\caption{Ablation study results on HotpotQA dev set. PR(0,1) stands for giving 0 score to non-gold documents and 1 score to all gold documents when preparing pairwise labels, and PR(0,1,2) stands for giving 2 score to the gold document with answer span.
}
\resizebox{1.0\columnwidth}{!}{
\begin{tabular}{l|c|c|c|c|c}
\hline
& EM\textsubscript{S} & Recall\textsubscript{S} & Acc\textsubscript{span} & joint EM & joint $F_1$ \\ \hline
BERT only & 70.65 & 89.16 & 90.08 & 31.87 & 59.33 \\ \hline
~~+MHSA & 87.07 & 94.65 & 92.54 & 38.54 & 65.00 \\ \hline
~~~~+PR(0,1) & 89.76 & 94.75 & 94.53 & 39.53 & 65.44 \\ \hline
~~~~+PR(0,1,2) & 91.40 & 95.61 & 95.86 & 39.89 & 66.45 \\ \hline
\end{tabular}}
\label{tab:doc_ablation}
\end{table}

We only present the ablation studies on our SAE system using \textit{BERT base uncased} model. We first show the results of ablation studies on the document selection module in Table \ref{tab:doc_ablation}. ``EM\textsubscript{S}'' and ``Recalls\textsubscript{S}'' measure the accuracy and recall of both two gold documents are selected; ``Acc\textsubscript{span}'' measures the accuracy of the gold document with answer span being selected. The ``BERT only'' model, similar with the document selection model in \cite{xiao2019dfgn}, is used as baseline. The MHSA mechanism, which allows information from different documents to interact with each other, brings the most gain over the baseline in terms of all measurements. The pairwise learning-to-rank loss function, which allows documents to be compared directly against each other and provides finer-grained information for the model to learn, further improves the performance. Further, scoring documents that contain the answer span to 2 is better than weighting the two gold documents equally. Finally, we can see that better document filter always results in better answer and support sentence prediction performance.


\begin{table}[]
\centering
\caption{Ablation study results on HotpotQA dev set.}
\resizebox{0.65\columnwidth}{!}{
\begin{tabular}{l|c|c}
\hline
& joint EM & joint $F_1$ \\ \hline
full model & 39.89 & 66.45 \\ \hline \hline
~~~-mixed attn & 39.59 & 66.28 \\ \hline
~~~-attn sum & 38.04 & 65.33 \\ \hline \hline
~~~-GNN & 38.46 & 65.53 \\ \hline
~~~-type 1 edge & 38.15 & 65.00 \\ \hline
~~~-type 2 edge & 39.55 & 66.13 \\ \hline
~~~-type 3 edge & 39.32 & 66.03 \\ \hline
~~~-type 2\&3 edge & 39.16 & 65.76 \\ \hline
\end{tabular}}
\label{tab:ablation}
\end{table}

Then we ablate different components in the ``Answer and Explain'' module, and we measure the model performance in these circumstances in terms of joint EM and $F_1$. From Table \ref{tab:ablation} it can be observed:
\begin{itemize}
\item If only self-attention with span logits when calculating attention weights for sentence attentive pooling (``-mixed attn'') is used, the performance drops marginally, indicating that the mixed attention mechanism helps but in a limited extent. However, replacing attention based sentence summarization with simple averaging (``-attn sum'') deteriorates the performance a lot (the joint EM drops almost 2\%), which proves the effectiveness of attention based sentence summarization. 
\item For the GNN module, if we do not use GNN based message passing over sentence embeddings, the performance degrades by $\sim$1.4\% for EM and $\sim$0.9\% for $F_1$. If we remove one of the three types of edges, the results show that the first type of edges, i.e. the connection among sentences within the same documents, contributes the most while other two types of edges contribute marginally. But when we remove both of them, the results further drops, indicating the usefulness of cross-document information.
\end{itemize}

\subsection{Result analysis}

In order to better understand the performance of the proposed SAE system, analysis is done based on different reasoning types in the development set. The reasoning type of an example is provided as part of the annotation by the data set. It has two categories: ``bridge'' and ``comparison''. The ``bridge'' type of reasoning requires the model to be able to find a bridge entity before reaching the final answer, while the ``comparison'' type of reasoning requires the model to compare the attributes of two entities and then give the answer. 
We calculate the joint EM and $F_1$ in each categorization (full sets of performance measurements is put in supplementary materials). We compare our proposed system with the baseline model and the DFGN model \cite{xiao2019dfgn}\footnote{only this model's output is available online} under these two reasoning types. 

In Table \ref{tab:reason}, we show the performance comparison in terms of joint EM and $F_1$ score under the ``bridge'' and ``comparison'' reasoning types. Our proposed SAE system delivers better performance under both reasoning types, and improvement over the DFGN model on the ``bridge'' type of reasoning is bigger than that of the "comparison" type. This proves our SAE system is better at dealing with bridging type of reasoning. Another interesting finding is that for all models the performance under ``comparison'' type is higher than ``bridge'' type. For our model, we conjecture that this is due to the bridge entity may be far from the question entities in context, thus is hard to find it. However, attributes of entities are usually close and are less possible to be ignored for ``comparison'' type of reasoning.

\begin{figure}[t]
    \centering
    \includegraphics[width=1.0\linewidth]{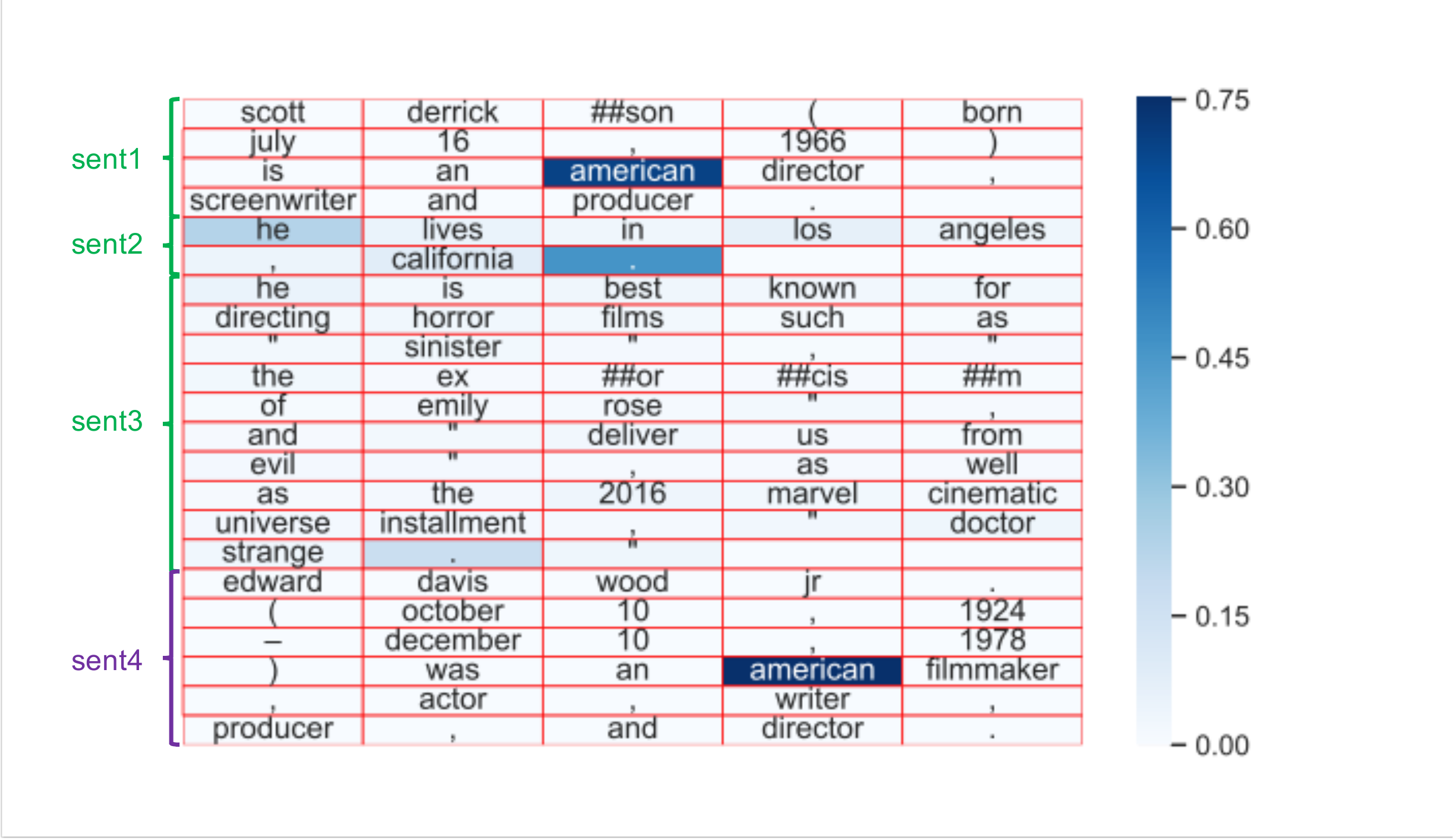}
    \caption{Attention heatmap of a sample from dev set. Each cell is a word piece token returned by BERT. Sentences with different colors are from different documents.}
    \label{fig:attention}
\end{figure}

In Figure \ref{fig:attention} we also show the sentence summarization attention heatmap of a sample in dev set. The question of this sample is ``Were Scott Derrickson and Ed Wood of the same nationality?''. It clearly shows that the model attends to supporting sentences with the word ``american'', which indicates the nationality of both ``Scott Derrickson'' and ``Ed Wood''. 
We include more ablation study results and analysis in supplementary materials.

\begin{table}[]
\caption{Performance comparison in terms of joint EM and $F_1$ scores under different reasoning types.}
\begin{center}
\resizebox{1.0\columnwidth}{!}{
\begin{tabular}{c|c|c|c|c}
\hline
\multirow{2}{*}{} & \multicolumn{2}{c|}{Bridge (5918 samples)} & \multicolumn{2}{c}{Comparison (1487 samples)} \\ \cline{2-5} 
 & joint EM & joint $F_1$ & joint EM & joint $F_1$ \\ \hline
Baseline & 8.80 & 39.77 & 20.91 & 43.24 \\ \hline
DFGN & 30.09 & 58.61 & 47.95 & 64.79 \\ \hline
SAE & 37.07 & 66.12 & 51.18 & 67.73 \\ \hline
\end{tabular}}
\end{center}
\label{tab:reason}
\end{table}

\section{Conclusion}
We propose a new effective and interpretable system to tackle the multi-hop RC problem over multiple documents. Our system first accurately filters out unrelated documents and then performs joint prediction of answer and supporting evidence. Several novel ideas to train the document filter model and the model for answer and support sentence prediction are presented. Our proposed system attains competitive results on the HotpotQA blind test set compared to existing systems. We would like to thank Peng Qi of Stanford University for running evaluation on our submitted models. This work is partially supported by Beijing Academy of Artificial Intelligence (BAAI).

{\small
\bibliography{hotpotqa}

\begin{thebibliography}{}

\bibitem[\protect\citeauthoryear{Cao, Fang, and Tao}{2019}]{cao2019bag}
Cao, Y.; Fang, M.; and Tao, D.
\newblock 2019.
\newblock Bag: Bi-directional attention entity graph convolutional network for
  multi-hop reasoning question answering.
\newblock In {\em Proceedings of the 2019 Conference of the North American
  Chapter of the Association for Computational Linguistics: Human Language
  Technologies, Volume 1 (Long and Short Papers)},  357--362.

\bibitem[\protect\citeauthoryear{Chen and
  Durrett}{2019}]{chen2019understanding}
Chen, J., and Durrett, G.
\newblock 2019.
\newblock Understanding dataset design choices for multi-hop reasoning.
\newblock In {\em Proceedings of the 2019 Conference of the North American
  Chapter of the Association for Computational Linguistics: Human Language
  Technologies, Volume 1 (Long and Short Papers)},  4026--4032.

\bibitem[\protect\citeauthoryear{Clark and Gardner}{2018}]{clark2018simple}
Clark, C., and Gardner, M.
\newblock 2018.
\newblock Simple and effective multi-paragraph reading comprehension.
\newblock In {\em Proceedings of the 56th Annual Meeting of the Association for
  Computational Linguistics (Volume 1: Long Papers)},  845--855.

\bibitem[\protect\citeauthoryear{De~Cao, Aziz, and
  Titov}{2019}]{de2018question}
De~Cao, N.; Aziz, W.; and Titov, I.
\newblock 2019.
\newblock Question answering by reasoning across documents with graph
  convolutional networks.
\newblock In {\em Proceedings of the 2019 Conference of the North American
  Chapter of the Association for Computational Linguistics: Human Language
  Technologies, Volume 1 (Long and Short Papers)},  2306--2317.

\bibitem[\protect\citeauthoryear{Devlin \bgroup et al\mbox.\egroup
  }{2019}]{devlin2018bert}
Devlin, J.; Chang, M.-W.; Lee, K.; and Toutanova, K.
\newblock 2019.
\newblock Bert: Pre-training of deep bidirectional transformers for language
  understanding.
\newblock In {\em Proceedings of the 2019 Conference of the North American
  Chapter of the Association for Computational Linguistics: Human Language
  Technologies, Volume 1 (Long and Short Papers)},  4171--4186.

\bibitem[\protect\citeauthoryear{Dhingra \bgroup et al\mbox.\egroup
  }{2018}]{dhingra2018neural}
Dhingra, B.; Jin, Q.; Yang, Z.; Cohen, W.; and Salakhutdinov, R.
\newblock 2018.
\newblock Neural models for reasoning over multiple mentions using coreference.
\newblock In {\em Proceedings of the 2018 Conference of the North American
  Chapter of the Association for Computational Linguistics: Human Language
  Technologies, Volume 2 (Short Papers)}, volume~2,  42--48.

\bibitem[\protect\citeauthoryear{Ko{\v{c}}isk{\`y} \bgroup et al\mbox.\egroup
  }{2018}]{kovcisky2018narrativeqa}
Ko{\v{c}}isk{\`y}, T.; Schwarz, J.; Blunsom, P.; Dyer, C.; Hermann, K.~M.;
  Melis, G.; and Grefenstette, E.
\newblock 2018.
\newblock The narrativeqa reading comprehension challenge.
\newblock {\em Transactions of the Association of Computational Linguistics}
  6:317--328.

\bibitem[\protect\citeauthoryear{Liu and others}{2009}]{liu2009learning}
Liu, T.-Y., et~al.
\newblock 2009.
\newblock Learning to rank for information retrieval.
\newblock {\em Foundations and Trends{\textregistered} in Information
  Retrieval} 3(3):225--331.

\bibitem[\protect\citeauthoryear{Liu \bgroup et al\mbox.\egroup
  }{2019}]{liu2019roberta}
Liu, Y.; Ott, M.; Goyal, N.; Du, J.; Joshi, M.; Chen, D.; Levy, O.; Lewis, M.;
  Zettlemoyer, L.; and Stoyanov, V.
\newblock 2019.
\newblock Roberta: A robustly optimized bert pretraining approach.
\newblock {\em arXiv preprint arXiv:1907.11692}.

\bibitem[\protect\citeauthoryear{Min \bgroup et al\mbox.\egroup
  }{2018}]{min2018efficient}
Min, S.; Zhong, V.; Socher, R.; and Xiong, C.
\newblock 2018.
\newblock Efficient and robust question answering from minimal context over
  documents.
\newblock In {\em Proceedings of the 56th Annual Meeting of the Association for
  Computational Linguistics (Volume 1: Long Papers)},  1725--1735.

\bibitem[\protect\citeauthoryear{Nishida \bgroup et al\mbox.\egroup
  }{2019}]{nishida2019answering}
Nishida, K.; Nishida, K.; Nagata, M.; Otsuka, A.; Saito, I.; Asano, H.; and
  Tomita, J.
\newblock 2019.
\newblock Answering while summarizing: Multi-task learning for multi-hop qa
  with evidence extraction.
\newblock In {\em Proceedings of the 57th Annual Meeting of the Association for
  Computational Linguistics},  2335--2345.

\bibitem[\protect\citeauthoryear{Paszke \bgroup et al\mbox.\egroup
  }{2017}]{paszke2017automatic}
Paszke, A.; Gross, S.; Chintala, S.; Chanan, G.; Yang, E.; DeVito, Z.; Lin, Z.;
  Desmaison, A.; Antiga, L.; and Lerer, A.
\newblock 2017.
\newblock Automatic differentiation in pytorch.

\bibitem[\protect\citeauthoryear{Rajpurkar \bgroup et al\mbox.\egroup
  }{2016}]{rajpurkar2016squad}
Rajpurkar, P.; Zhang, J.; Lopyrev, K.; and Liang, P.
\newblock 2016.
\newblock Squad: 100,000+ questions for machine comprehension of text.
\newblock In {\em Proceedings of the 2016 Conference on Empirical Methods in
  Natural Language Processing},  2383--2392.

\bibitem[\protect\citeauthoryear{Rajpurkar, Jia, and
  Liang}{2018}]{rajpurkar2018know}
Rajpurkar, P.; Jia, R.; and Liang, P.
\newblock 2018.
\newblock Know what you don’t know: Unanswerable questions for squad.
\newblock In {\em Proceedings of the 56th Annual Meeting of the Association for
  Computational Linguistics (Volume 2: Short Papers)}, volume~2,  784--789.

\bibitem[\protect\citeauthoryear{Reddy, Chen, and
  Manning}{2019}]{reddy2018coqa}
Reddy, S.; Chen, D.; and Manning, C.~D.
\newblock 2019.
\newblock Coqa: A conversational question answering challenge.
\newblock {\em Transactions of the Association for Computational Linguistics}
  7:249--266.

\bibitem[\protect\citeauthoryear{Rei and S{\o}gaard}{2019}]{rei2019jointly}
Rei, M., and S{\o}gaard, A.
\newblock 2019.
\newblock Jointly learning to label sentences and tokens.
\newblock In {\em Proceedings of the AAAI Conference on Artificial
  Intelligence}, volume~33,  6916--6923.

\bibitem[\protect\citeauthoryear{Seo \bgroup et al\mbox.\egroup
  }{2016}]{seo2016bidirectional}
Seo, M.; Kembhavi, A.; Farhadi, A.; and Hajishirzi, H.
\newblock 2016.
\newblock Bidirectional attention flow for machine comprehension.
\newblock {\em arXiv preprint arXiv:1611.01603}.

\bibitem[\protect\citeauthoryear{Song \bgroup et al\mbox.\egroup
  }{2018}]{song2018exploring}
Song, L.; Wang, Z.; Yu, M.; Zhang, Y.; Florian, R.; and Gildea, D.
\newblock 2018.
\newblock Exploring graph-structured passage representation for multi-hop
  reading comprehension with graph neural networks.
\newblock {\em arXiv preprint arXiv:1809.02040}.

\bibitem[\protect\citeauthoryear{Tay \bgroup et al\mbox.\egroup
  }{2018}]{tay2018densely}
Tay, Y.; Luu, A.~T.; Hui, S.~C.; and Su, J.
\newblock 2018.
\newblock Densely connected attention propagation for reading comprehension.
\newblock In {\em Advances in Neural Information Processing Systems},
  4911--4922.

\bibitem[\protect\citeauthoryear{Tu \bgroup et al\mbox.\egroup
  }{2019}]{tu2019hdegraph}
Tu, M.; Wang, G.; Huang, J.; Tang, Y.; He, X.; and Zhou, B.
\newblock 2019.
\newblock Multi-hop reading comprehension across multiple documents by
  reasoning over heterogeneous graphs.
\newblock In {\em Proceedings of the 57th Annual Meeting of the Association for
  Computational Linguistics},  2704--2713.

\bibitem[\protect\citeauthoryear{Vaswani \bgroup et al\mbox.\egroup
  }{2017}]{vaswani2017attention}
Vaswani, A.; Shazeer, N.; Parmar, N.; Uszkoreit, J.; Jones, L.; Gomez, A.~N.;
  Kaiser, {\L}.; and Polosukhin, I.
\newblock 2017.
\newblock Attention is all you need.
\newblock In {\em Advances in neural information processing systems},
  5998--6008.

\bibitem[\protect\citeauthoryear{Wang \bgroup et al\mbox.\egroup
  }{2019}]{wang2019evidence}
Wang, H.; Yu, D.; Sun, K.; Chen, J.; Yu, D.; Roth, D.; and McAllester, D.
\newblock 2019.
\newblock Evidence sentence extraction for machine reading comprehension.
\newblock {\em arXiv preprint arXiv:1902.08852}.

\bibitem[\protect\citeauthoryear{Welbl, Stenetorp, and
  Riedel}{2018}]{welbl2018constructing}
Welbl, J.; Stenetorp, P.; and Riedel, S.
\newblock 2018.
\newblock Constructing datasets for multi-hop reading comprehension across
  documents.
\newblock {\em Transactions of the Association of Computational Linguistics}
  6:287--302.

\bibitem[\protect\citeauthoryear{Xiao \bgroup et al\mbox.\egroup
  }{2019}]{xiao2019dfgn}
Xiao, Y.; Qu, Y.; Qiu, L.; Zhou, H.; Li, L.; Zhang, W.; and Yu, Y.
\newblock 2019.
\newblock Dynamically fused graph network for multi-hop reasoning.
\newblock In {\em Proceedings of the 57th Annual Meeting of the Association for
  Computational Linguistics},  6140--6150.

\bibitem[\protect\citeauthoryear{Xiong, Zhong, and
  Socher}{2016}]{xiong2016dynamic}
Xiong, C.; Zhong, V.; and Socher, R.
\newblock 2016.
\newblock Dynamic coattention networks for question answering.
\newblock {\em arXiv preprint arXiv:1611.01604}.

\bibitem[\protect\citeauthoryear{Yang \bgroup et al\mbox.\egroup
  }{2018}]{yang2018hotpotqa}
Yang, Z.; Qi, P.; Zhang, S.; Bengio, Y.; Cohen, W.; Salakhutdinov, R.; and
  Manning, C.~D.
\newblock 2018.
\newblock Hotpotqa: A dataset for diverse, explainable multi-hop question
  answering.
\newblock In {\em Proceedings of the 2018 Conference on Empirical Methods in
  Natural Language Processing},  2369--2380.

\bibitem[\protect\citeauthoryear{Zhong \bgroup et al\mbox.\egroup
  }{2019}]{zhong2019coarse}
Zhong, V.; Xiong, C.; Keskar, N.~S.; and Socher, R.
\newblock 2019.
\newblock Coarse-grain fine-grain coattention network for multi-evidence
  question answering.
\newblock {\em arXiv preprint arXiv:1901.00603}.

\bibitem[\protect\citeauthoryear{Zhou, Huang, and
  Zhu}{2018}]{zhou2018interpretable}
Zhou, M.; Huang, M.; and Zhu, X.
\newblock 2018.
\newblock An interpretable reasoning network for multi-relation question
  answering.
\newblock In {\em Proceedings of the 27th International Conference on
  Computational Linguistics},  2010--2022.

\end{thebibliography}
\bibliographystyle{aaai}
}

\end{document}